\documentclass{article}

\usepackage{PRIMEarxiv}

\usepackage[utf8]{inputenc} 
\usepackage[T1]{fontenc}    
\usepackage{hyperref}       
\usepackage{url}            
\usepackage{booktabs}       
\usepackage{amsfonts}       
\usepackage{nicefrac}       
\usepackage{microtype}      
\usepackage{lipsum}
\usepackage{fancyhdr}       
\usepackage{graphicx}       
\graphicspath{{media/}}     

\usepackage{amsthm,amsmath,amssymb}

\title{SwinVRNN: A Data-Driven Ensemble Forecasting Model via Learned Distribution Perturbation
}

\author{
  Yuan Hu\thanks{These authors contributed equally to this work}, Lei Chen$^*$, Zhibin Wang, Hao Li\thanks{Corresponding author} \\
  DAMO Academy \\
  Alibaba Group \\
  Beijing, China\\
  \texttt{\{lavender.hy, fanjiang.cl, zhibin.waz, lihao.lh\}@alibaba-inc.com} \\
}

\begin{document}
\maketitle

\begin{abstract}
Data-driven approaches for medium-range weather forecasting 
are recently shown extraordinarily promising for ensemble forecasting for their fast inference speed compared to traditional numerical weather prediction (NWP) models, but their forecast accuracy can hardly match the state-of-the-art operational ECMWF Integrated Forecasting System (IFS) model.
Previous data-driven attempts achieve ensemble forecast using some simple perturbation methods, like initial condition perturbation and Monte Carlo dropout.
However, they mostly suffer unsatisfactory ensemble performance, which is arguably attributed to the sub-optimal ways of applying perturbation. 
We propose a Swin Transformer-based Variational Recurrent Neural Network (SwinVRNN), which is a stochastic weather forecasting model combining a SwinRNN predictor with a perturbation module.
SwinRNN is designed as a Swin Transformer-based recurrent neural network, which predicts future states deterministically.
Furthermore, to model the stochasticity in prediction, we design a perturbation module following the Variational Auto-Encoder paradigm to learn multivariate Gaussian distributions of a time-variant stochastic latent variable from data. 
Ensemble forecasting can be easily achieved by perturbing the model features leveraging noise sampled from the learned distribution.
We also compare four categories of perturbation methods for ensemble forecasting, 
i.e. fixed distribution perturbation, learned distribution perturbation, MC dropout, and multi model ensemble.
Comparisons on WeatherBench dataset show the learned distribution perturbation method using our SwinVRNN model achieves superior forecast accuracy and reasonable ensemble spread due to joint optimization of the two targets.
More notably, SwinVRNN surpasses operational IFS on surface variables of 2-m temperature and 6-hourly total precipitation at all lead times up to five days.
\end{abstract}


\section{Introduction}




Medium-range weather prediction is crucial to various weather-sensitive activities, such as agricultural operations, water resource management, public transportation, social activities, disaster prevention, etc.
Accurate weather forecasts can prevent massive loss of life and property from extreme weather events like heatwaves, tropical cyclones, and floods.

Current operational systems for medium-range weather forecast generally adopt the Numerical Weather Prediction (NWP) models,
which build on fluid dynamic equations and good estimation of the initial conditions including state variables of the atmosphere, land and ocean. 
Though very successful, the NWP approaches suffer from the issues of expensive computation and model errors introduced by various physics parameterization schemes \cite{fourcasenet}. 
Recently, the limitations in the NWP approaches motivate increasing interest of researchers in purely data-driven weather forecasting approaches. 
One of the most significant advantages of the data-driven Deep Learning (DL) models is their orders of magnitude faster inference (i.e. prediction) speed, which can enable extremely large ensemble forecasts. 


Many attempts \cite{weatherbench, resnet-paper, cube-paper, fourcasenet} have been made to utilize DL-based data-driven models for weather prediction.
Among these attempts, \cite{weatherbench} proposed a dataset, i.e. WeatherBench, to provide a common benchmark for evaluating different data-driven models, and adopted a simple convolutional neural network (CNN) to set a basic baseline on WeatherBench;
\cite{resnet-paper} trained a ResNet \cite{resnet} model to predict geopotential, temperature and precipitation at $5.625^{\circ}$ on WeatherBench following \cite{weatherbench}, and especially, they pre-trained the model using historical climate model simulation, which helps to prevent overfitting and gives better testing scores;
\cite{cube-paper} first projected global data onto a cubed sphere, minimizing the distortion on the cube faces, and then used U-Net \cite{unet} to perform prediction on each cube face. 
These methods have achieved some improvements, but their performance gap with the Integrated Forecasting System (IFS), the state-of-the-art NWP method so far, is still apparent.

FourCastNet \cite{fourcasenet} has been the first method that can achieve comparable performance with IFS in small-scale variables including surface wind speed and precipitation at $0.25^{\circ}$ resolution.
It uses a Fourier-based neural network \cite{afno} with a vision transformer (VIT) \cite{vit} backbone to generate global forecasts iteratively. 
With rapid inference speed, 
FourCastNet can generate ensemble forecasts by perturbing the initial condition using Gaussian random noise.
According to the experiments in  \cite{fourcasenet}, noticeable improvements can be observed  over the unperturbed control for the 100-member FourCastNet ensemble.
This demonstrates the great potential of data-driven ensemble forecasting.
However, we argue that the perturbation approach adopted by FourCastNet is not an optimal one, where the noise is sampled from a fixed standard normal distribution.
This means it cannot adjust the perturbation intensity for different spatial locations, such as the sharply fluctuating area in the mid-latitude zone of Z500 field or the gently changing area in the low-latitude region of Z500 field, or for different atmospheric variables, such as smooth geopotential field Z500 and sharper temperature field T850.
Actually, we also implement the fixed distribution perturbation method in this paper, and find that its ensemble spread is either extremely limited or unreasonably explosive.
There are also some other perturbation methods, such as Monte-Carlo dropout (MC dropout), parametric prediction, and categorical prediction, which have been utilized in \cite{probability} for probabilistic weather forecasting, but none of the models is able to match the skill of the operational IFS model.


Most of the existing data-driven methods apply deterministic models to forecast future states.
However, due to the inherent uncertainty in the dynamics of the atmosphere, a historical state likely leads to a number of future states within some range.
Due to the stochastic nature of medium-range weather forecasting, the models trained with deterministic loss can hardly achieve optimal future weather forecasts. 
For example, the expected value of all possible outcomes will be predicted if mean squared error loss is used.

Based on above observations, in this paper, we propose a Swin Transformer-based Variational Recurrent Neural Network (SwinVRNN) model that combines a deterministic SwinRNN predictor with a perturbation module. 
For SwinRNN, we apply an encoder-decoder architecture with Swin Transformer blocks \cite{swintransformer} as the basic building blocks.
SwinRNN itself is a deterministic predictor that forecasts future state variables deterministically.
The perturbation module is proposed to learn a stochastic latent variable, which is used to model the uncertainty in prediction at each time step. 
Specifically, it learns the posterior and prior Gaussian distributions of the latent variables following the Variational Auto-Encoder paradigm.
Then it enables efficient and effective ensemble forecasting by perturbing the SwinRNN features by leveraging the learned noise.
It is worth noting that the overall objective of the model is composed of the reconstruction loss between forecasts and truth and the KL divergence loss between posterior and prior distributions, where the weight of the second loss can control the diversity of multiple different predictions. 
Training with this objective can be viewed as equivalent to optimizing the forecast accuracy as well as the ensemble spread at the same time.

According to different perturbation types including input perturbation, model parameter perturbation, feature perturbation, and model perturbation, we design four ensemble forecasting approaches based on the proposed SwinRNN and SwinVRNN models, namely fixed distribution perturbation method, MC dropout, learned distribution perturbation method, and multi-model ensemble.
Comprehensive experiments on WeatherBench dataset demonstrate that the learned distribution perturbation implemented by SwinVRNN yields significantly better ensemble results than those generated by the fixed distribution perturbation method and MC dropout, in terms of the forecast accuracy (measured by RMSE) and ensemble spread.
In addition, multi-model ensemble of SwinVRNN is able to further boost the ensemble performance.
Visualization of ensemble forecasts on several specific locations in Z500 field shows that the forecasts generated by our models have better consistency with the ERA5 truth on the overall tendency as well as more reasonable ensemble spread around the truth.
We attribute these good results to the ability of our model to adjust the perturbation intensity dynamically according to the variation of atmospheric variables in the temporal and spatial dimension. 
More notably, the forecast accuracy of the purely data-driven based SwinVRNN model surpasses that of IFS in 2-m temperature (T2M), and 6-hourly total precipitation (TP) at all lead times up to five days and approaches performance of IFS on $850$ hPa temperature (T850) within five days and finally surpasses it at the lead time of five days.

The contributions of this paper are three-fold, summarized as follows:
\begin{enumerate}
\item To the best of our knowledge, SwinVRNN is the first to apply a stochastic recurrent deep learning model to perform medium-range ensemble weather forecasting.

\item We make a comprehensive investigation and discussion on the forecast accuracy and ensemble spread of different data-driven ensemble forecasting methods, including fixed distribution perturbation, learned distribution perturbation, model parameters perturbation, and multi-model ensemble.


\item SwinVRNN achieves a new state-of-the-art among data-driven weather forecasting models, and more notably, it outperforms IFS in variables including T2M and TP at all lead times up to five days on WeatherBench dataset at $5.625^\circ$ spatial resolution.
\end{enumerate}





\section{Methods}

\begin{figure}[t]
\centering
\includegraphics[width=16cm]{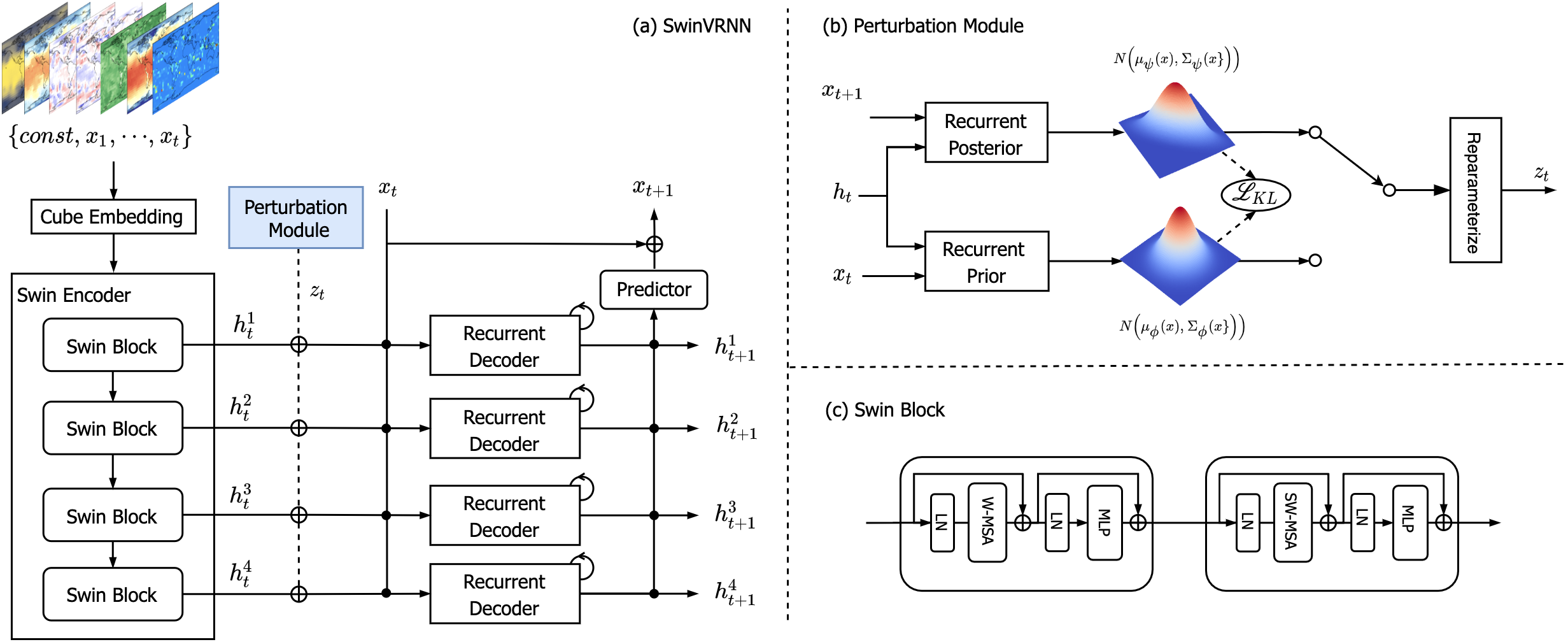}
\caption{Overall architecture of SwinVRNN. (a) The SwinVRNN model is composed of the SwinRNN backbone and a perturbation module. The SwinRNN backbone consists of four components, i.e. a cube embedding block, a Swin encoder, multi-scale recurrent decoders, and a predictor. The cube embedding block takes as input the historical frames (including $\{x_1,\cdots,x_t\}$ and two constant fields), and encodes them as a series of tokens.
The Swin encoder consists of four stages and extracts different scales of hidden features from them. Each recurrent decoder accepts a corresponding hidden feature $h_t^i$ and the original signal $x_t$ as inputs and updates the $h_{t+1}^i$ based on their relation.
Finally, the predictor forecasts the variables' change between two consecutive time steps from the multi-scale features, which is then added to $x_t$ to get the final output $x_{t+1}$. 
(b) The perturbation module is composed of a recurrent posterior network and a recurrent prior network. The posterior network takes as input $x_{t+1}$ and $h_t$ and outputs the posterior Gaussian distribution of a random latent variable $z_t$. 
The prior network takes as input $x_t$ and $h_t$ and outputs the prior Gaussian distribution of $z_t$. KL divergence loss is applied between the two learned distributions. The switch means that $z_t$ is sampled from the posterior during training phase and prior during inference. After $z_t$ is obtained, it is added to each scale of $h_t$ to perturb the hidden features as a noise, as shown in (a). (c) The architecture of Swin block, which is used in the Swin encoder, recurrent decoder, and posterior and prior networks.
\label{fig-swinvrnn}}
\end{figure}

We formulate the medium-range weather prediction problem as a spatio-temporal sequence prediction task. 
Suppose $x_{1:T}=(x_1, x_2, \cdots, x_T)$ denotes a sequence of atmospheric state variables, each $x_t$ representing the state variables at time step $t$.
Assume we have historical observations $x_{1:t}$.
Our goal is to forecast future states $x_{t+1:T}$. 
The projection is achieved through a prediction model $p_\theta(x_{t+1:T}|x_{1:t})$.
We implement $p_\theta(\cdot)$ with the SwinRNN model which is a deterministic recurrent network that predicts future states iteratively. 
In addition, a perturbation module is proposed to learn a latent variable $z_t$, which is then incorporated into the SwinRNN model. 
For this perturbation module we apply the Variational Auto-Encoder paradigm.
With these designs, we develop a Swin transformer based Variational Recurrent Neural Network (SwinVRNN). 
The SwinVRNN is a stochastic model with $z_t$ to model the stochastic phenomena during each time step prediction. 
In this way, we re-express the prediction model as $p_\theta(x_{t+1:T}|x_{1:t}, z_t)$, parameterized by $\theta$.
Figure \ref{fig-swinvrnn}
illustrates the overall architecture of the proposed SwinVRNN.


\subsection{Deterministic SwinRNN} \label{sec-swinrnn}


SwinRNN consists of four components: a cube embedding block, a Swin encoder, multi-scale recurrent decoders, and a predictor. The cube embedding block takes as input the historical variable states of different pressure levels as well as two constant variables,  detailed in Table \ref{table-variables}.
The input variables make a tensor with shape $C_{in} \times T \times H \times W$, where $C_{in}$ denotes the number of atmospheric variables, $T$ denotes history time steps, and $H$, $W$ denote the image spatial size.
A 3D convolution operation with both kernel and stride being $T \times 1 \times 1$ is applied to transform this tensor to a new one with shape $C_{out} \times 1 \times H \times W$, which is further rearranged into a sequence of $HW$ tokens where each token is of $C_{out}$ channels.

The Swin encoder is a Swin Transformer based architecture.
Swin Transformer \cite{swintransformer} is a variant of the vision transformer, which limits self-attention computation to non-overlapping local windows.
Such a shifted window scheme greatly reduces the computational complexity and meanwhile enables efficient interaction between different windows. 
Following \cite{swintransformer}, the Swin encoder consists of four stages, among which each of the first three stages is followed by $2 \times$ spatial downsampling operations (similar to the upscaling operation in NWP) to form a hierarchical architecture.
Each stage is composed of two successive Swin Transformer blocks, as shown in Figure \ref{fig-swinvrnn} (c). Four scale features $(h_t^1, h_t^2, h_t^3, h_t^4)$ are extracted to initialize the hidden states at time step $t$, where $h_t^1$ has the same spatial size with the input, and $h_t^4$ is $\frac{1}{8}$ of the original size of the input.

The multi scale hidden states are updated by separate recurrent decoders. Each decoder takes as input the current variable state $x_t$ and a hidden state $h_t$, and then six consecutive Swin blocks are utilized to model their relations and update the hidden state to get $h_{t+1}$.
Specifically, we first fuse the input and hidden states by concatenation and linear projection, 
and then pass the fusion results to the Swin blocks, in which self attention models the relation between the current input and the historical hidden feature and updates the hidden states accordingly. 
Thanks to the shifted window design, both local and global information are injected to the state feature. Window size is set to 8 for all scales. 
In this way, we obtain $(h_{t+1}^1, h_{t+1}^2, h_{t+1}^3, h_{t+1}^4)$ for next time step prediction. Finally, the predictor aggregates the updated multi scale hidden states and predicts the change of each atmosphere variable from $t$ to $t+1$, which is then added to the input $x_t$ to produce the next step prediction $x_{t+1}$.
The residual prediction scheme is somewhat consistent with the way NWP simulation integrates in the temporal dimension.

Note that the proposed SwinRNN model differs from previous convolutional recurrent neural networks (RNNs), such as ConvLSTM \cite{convlstm}, ConvGRU \cite{convgru}, etc., in three aspects.
1) Recurrent design: SwinRNN extracts features from all the historical frames at once, and its auto-regressive prediction starts from the first time step of a future sequence, which is more memory-efficient compared with recurring from the initial frame in other convolutional RNNs.
2) Memory update: Previous convolutional RNNs \cite{convlstm, convgru, predrnn} adopt different gates to update memory states, while our SwinRNN leverages self-attention operation to achieve the interaction between current input and history information so as to update the hidden states. 
3) Multi-scale prediction: Each scale in SwinRNN maintains its own hidden state, and each time step prediction is performed by fusing multi-scale features, which is similar to the multi-scale design of \cite{nature}.
In contrary to single-scale prediction in previous convolutional RNNs, our SwinRNN uses features in different resolutions to forecast atmosphere events with different spatial scales, such as planetary-scale atmospheric circulation, and synoptic-scale mid and high-latitude troughs and ridges, etc.

\subsection{Perturbation Module} \label{sec-perturbation-module}
Since SwinRNN is a deterministic prediction model, we resort to a perturbation module to model the stochasticity in future state prediction at each time step.
The perturbation module predicts a multivariate Gaussian distribution of a random variable $z_t$ at each time. 
The random sampled $z_t$ plays a role of learned noise to perturb the hidden state features $h_t$. 
In this way, every time we sample a $z_t$, we obtain different predictions.

Assume $z_t$ is subjected to a prior distribution $p(z_t)$.
At each time step we aim to estimate the log-likelihood of the function $p(x_{t+1}|x_{1:t}, z_t)=\int_{z_t} p_\theta(x_{t+1}|x_{1:t}, z_t) p(z_t)dz_t$, which is intractable due to the unknown true prior distribution over $z_t$.
A typical solution is to use an encoder to infer the posterior $p(z_t|x_{1:t+1})$ from the data $x_{1:t+1}$, and then a decoder to predict $x_{t+1} \sim p_\theta(x_{t+1}|x_{1:t}, z_t)$.
In practice, we adopt a Swin block based recurrent network  $q_\phi(z_t|x_{1:t+1})$ to approximate the true posterior by predicting the mean and covariance matrix of a conditional multivariate Gaussian distribution
$\mathcal{N}(\mu_\phi(x_{1:t+1}), \Sigma_\phi(x_{1:t+1})))$, where $\phi$ denotes the parameters of the posterior inference network.

To learn the parameters of the SwinRNN and the posterior network, we optimize the variational lower bound, as in the Stochastic Video Generation (SVG) \cite{svg}:
\begin{equation}
    \label{eq-L}
    \mathcal{L}_{\theta, \phi} =
    \mathbb{E}_{q_\phi(z_t|x_{1:t+1})} \log p_\theta(x_{t+1}|x_{1:t}, z_t)
    -
    \beta KL(q_\phi(z_t|x_{1:t+1}) \| p(z_t)).
\end{equation}

The expectation term in Equation (\ref{eq-L}) represents the reconstruction loss between the predicted and ground truth future states. MSE loss is used in practice. The second term denotes the KL divergence between the approximated posterior and the prior of the latent variable.
Note that the posterior $q_\phi(\cdot)$ takes as input the target states $x_{t+1}$ besides previous states $x_{1:t}$, which encourages $z_t$ to capture additional information that does not exist in previous states but is beneficial to the reconstruction loss. 
Also note that $q_\phi(\cdot)$ accepts $x_{1:t}$ as input due to the recurrent architecture of the posterior network.
However, the KL term prohibits the $z_t$ from simply copying the encoding of the $x_{t+1}$ by constraining the $q_\phi(z_t|x_{1:t+1})$ to be close to the prior distribution $p(z_t)$. 
The weighting factor $\beta$ can influence the prediction error and the diversity of the results by controlling the extent of the posterior being close to the prior. This means that the forecast accuracy and the ensemble spread are joint optimized through the objective, which are exactly the two concerned aspects for ensemble forecast. In this paper, we set $\beta$ equals $1e-4$ to achieve decent prediction accuracy as well as plausible diversity.

As for the prior distribution $p(z_t)$, previous methods usually assume it as a fixed Gaussian $\mathcal{N}(\mathbf{0}, \mathbf{I})$ \cite{sv2p} or learn the mean $\mu_\psi(x_{1:t})$ and standard deviation $\log \sigma_\psi(x_{1:t})$ of a conditional Gaussian distribution $\mathcal{N}(\mu_\psi(x_{1:t}), \sigma_\psi(x_{1:t}))$ from the history data $x_{1:t}$ \cite{svg, nuq}. 
Both of them are based on an assumption that their covariance matrix of the multivariate normal distribution is a diagonal one, where each entry of $z_t$ is mutually independent of the other.
However, this assumption is not reasonable when $z_t$ has a spatial dimension. 
According to Tobler's first law of geography that  ``all things are related, but nearby things are more related than distant things" \cite{geo-law}, random variables at different spatial locations have properties of autocorrelation. 
In this way, the covariance between  $z_t^{i,j}$ at two different spatial locations should be considered.
We then use another recurrent neural network to learn the mean $\mu_\psi(x_{1:t})$ and the covariance matrix $\Sigma_\psi(x_{1:t})$ of the approximated multivariate prior, where $\psi$ denotes the parameters of the prior inference network.

Re-parametrization trick \cite{vae} of multivariate normal distribution is used to permit effective optimization of Equation (\ref{eq-L}) as follows:
\begin{align}
  z_t = \mu &+ L \times \epsilon \nonumber \\ 
  LL^T &= \Sigma,  \\
  \epsilon \sim \mathcal{N} & (\mathbf{0}, \mathbf{I}) \nonumber
\end{align}
where $\mu$ and $\Sigma$ represent $\mu_\phi(x_{1:t+1})/\Sigma_\phi(x_{1:t+1})$ and $\mu_\psi(x_{1:t})/\Sigma_\psi(x_{1:t})$ estimated by the  posterior and prior network, respectively. $L$ is the Cholesky factor of $\Sigma$.

Figure \ref{fig-swinvrnn} (b) gives an illustration of the perturbation module. The posterior network takes as input the target variable $x_{t+1}$ and the hidden state $h_t$ and outputs the mean and covariance matrix of the posterior distribution. The prior network is similar except that it accepts $x_t$ as input instead of $x_{t+1}$.
The switch in Figure \ref{fig-swinvrnn} (b) represents that $z_t$ is resampled from the posterior during training and prior during inference, since $x_{t+1}$ is not available during inference. The architectures of the posterior network and the prior network are completely the same. Both of them are composed of a Swin block, as shown in Figure \ref{fig-swinvrnn} (c). After we get $z_t$, it is added to each scale of $h_t$ to perturb the hidden features as denoted by the dashed line in Figure \ref{fig-swinvrnn} (a). Forecasting is then performed based on the perturbed features.

\subsection{Training} \label{sec-training}
We adopt a two-phase strategy for training the proposed SwinVRNN. In the first phase, we disable the perturbation module and train the SwinRNN backbone individually. A scheduled sampling strategy is applied to bridge the discrepancy between training and inference. Specifically, exponential decay is adopted to slowly decrease the teacher forcing ratio from $1.0$ to $0.0$; that is, the probability of performing teacher forcing along the lead time dimension decays gradually. 
In the second phase, the whole network including the SwinRNN and the perturbation module is trained jointly. Several tips are applied to facilitate efficient fine tuning. First, we initialize the prior distribution as a standard normal distribution $\mathcal{N}(\mathbf{0}, \mathbf{I})$. Second, we apply a multiplier to $z_t$ before it is added to the $h_t$. The coefficient is initialized to zero and slowly increased to one, which facilitates the transition from the first phase to the second. Third, we apply a much smaller learning rate for the pretrained SwinRNN backbone, and a larger learning rate for the newly added perturbation module. We have empirically found that SwinVRNN can achieve desired performance with the two-phase training procedure.

As for the detailed experimental setup, in the first phase, the SwinRNN is trained for $100$ epoch using a cosine learning rate policy with an initial learning rate that equals $0.0002$. In the second phase, the whole model is finetuned for another $100$ epoch using the cosine policy. The initial learning rates for SwinRNN backbone and perturbation module are $0.00002$ and $0.0002$ respectively.
All experiments are optimized by AdamW using PyTorch and performed using 8 NVIDIA Tesla V100(32G) GPUs.



\subsection{Ensemble Forecast} \label{sec-ensemble-forecast}

It is intuitive and straightforward to generate ensemble forecasts using the proposed SwinVRNN model.
During inference, the prior network is utilized to predict the multivariate Gaussian distribution, from which $z_t$ is sampled for each future time step $t$. Then, the $z_t$ is added to $h_t$ to perturb the hidden features of the model.
Ensemble forecasts can be easily achieved by sampling the latent variable $z$ multiple times and computing the ensemble mean of the perturbed forecasts at each time step.
Ensemble forecasting is fast using SwinVRNN, i.e. 44 seconds for 100-member forecasting using a single NVIDIA Tesla V100 GPU.
We classify this ensemble method as the learned distribution perturbation method, since the perturbation noises are sampled from the learned distribution.

The proposed learned distribution perturbation method differs from the fixed distribution perturbation method adopted by FourCastNet \cite{fourcasenet} in three aspects. 
First, FourCastNet applies the perturbation on the input variables, while SwinVRNN applies it on the hidden features of the deep model. 
Second, the noises are drawn from the predefined $\mathcal{N}(\mathbf{0}, \mathbf{I})$ in FourCastNet, whereas in our model, the distributions are learned from the data, where the perturbation intensity can be adjusted dynamically for different spatial locations and channels. Third, FourCastNet assumes the diagonal covariance matrix of the noise regardless of the spatial correlation, while our model takes the covariance into account.
Comprehensive experiments are conducted in Section \ref{sec-ablation-ensemble} to compare the ensemble performance of learned distribution and fixed distribution perturbation methods, as well as MC dropout and multi-model ensemble methods. Experiments in Section \ref{sec-ablation-swinvrnn} also ablate the effectiveness of learning the covariance matrix.

\section{Experiments}

\subsection{Data}

\begin{table}[t]
 \caption{Atmospheric Variables Adopted to Train the Models in This Paper.}
 \label{table-variables}
 \centering
     \begin{tabular}{l c | l c | l c}
 \toprule
 3D Variables & Levels & 2D Variables & Levels & Constants & Levels \\
 \midrule
 Geopotential & $13$ & 2m\_temperature & $1$ & land\_binary\_mask & $1$ \\
 Temperature & $13$  &  10m\_u\_component\_of\_wind & $1$ & orography & $1$ \\
 Relative\_humidity & $13$ &  10m\_v\_component\_of\_wind & $1$ \\
 u\_component\_of\_wind & $13$ &  total\_precipitation & $1$ \\
 v\_component\_of\_wind & $13$ \\
 \bottomrule
 \end{tabular}
 \end{table}
 
WeatherBench dataset \cite{weatherbench} is used in this paper to evaluate the proposed method. Variables in WeatherBench are selected from the ERA5 reanalysis data \cite{era5}. 
In this paper, we choose 500 hPa geopotential (Z500), 850 hPa temperature (T850), 2-m temperature (T2M), and total 6-hourly accumulated precipitation (TP) as targets.
Considering the complex interactions between different atmospheric variables, we also choose several additional variables to provide context information for the target prediction, including geopotential, temperature, relative humidity, longitude-direction wind, and latitude-direction wind at all $13$ vertical layers, four single-layer fields ($2$m temperature, $10$m wind, and total precipitation), and two constant fields (land-sea mask and orography), which amounts to $71$ fields, as shown in Table \ref{table-variables}.

 WeatherBench contains $40$-years data from $1979$ to $2018$. Data from $2017$ to $2018$ are chosen as test set, and others are used for training. Three different resolutions are provided by WeatherBench, including $5.625^\circ$, $2.8125^\circ$, and $1.40525^\circ$. In this paper, $5.625^\circ$ is adopted to train the proposed model following \cite{weatherbench, resnet-paper}. In addition, 5-days forecasts are predicted conditioned on historical $36$-hours input, which results in a sequence consists of $26$ frames with temporal interval of $6$ hours.
In this way, the input variable states are represented as a 4D tensor with shape $71 \times 6 \times 32 \times 64$, which is encoded by the cube embedding block and then fed into the Swin encoder to extract multi-scale features. At each time step, the predictor forecasts 6 hour change of all time-variant input variables except $2$ constants, which results in a 4D residual tensor with shape $68 \times 20 \times 32 \times 64$. Recurrent predicting is achieved by every time forwarding the output of last time step back to the Swin decoder.

\subsection{Baselines}
We compare the proposed SwinVRNN model with three categories of baselines, namely weekly climatology, NWP baselines and data-driven baselines. 
The weekly climatology is a climatological forecast computed for each of the $52$ calendar weeks between $1979–2016$.
As for NWP baselines, we choose three NWP models provided by WeatherBench \cite{weatherbench}:
1) IFS (Integrated Forecast System) model, which is the state-of-the-art operational NWP model of the European Center for Medium-range Weather Forecasting (ECMWF), and runs at $9$ km horizontal resolution with $137$ vertical levels;
2) T63 model, which is the same model as IFS, but runs at $210$ km ($\sim$ $1.9^\circ$) with $62$ vertical levels;
3) T42 model, which runs at much coarser resolution ($310$ km or $\sim$ $2.8^\circ$) with $62$ vertical levels.
In addition, we also compare our model with prior data-driven forecasting models, which are listed as follows.
 
 
 \begin{enumerate}
     \item Na\"ive CNN \cite{weatherbench}. This is a very simple fully convolutional neural network (CNN) with five layers, which only takes as input two channels (i.e. Z500 and T850 at $5.625^\circ$) and forecasts their future states up to 5 days directly.
     \item Cubed UNet \cite{cube-paper}. Atmospheric variables in latitude-longitude coordinates are fistly mapped to the cubed-sphere grid off-line, and then a U-Net model \cite{unet} is performed individually on each cube face. The model predicts four fields: Z500, Z1000, T2M, and $\tau_{300-700}$ (300- to 700-hPa geopotential thickness) at $1.9^\circ$. The input fields consist of the above four variables and three additional prescribed fields.
     \item ResNet (pretrained) \cite{resnet-paper}. A deep ResNet \cite{resnet} with $19$ residual blocks is adopted to forecast four atmospheric variables, including Z500, T850, T2M, and TP at $5.625^\circ$. In addition, the model is firstly pretrained using historical climate model output to improve forecast skill. Several variables at seven vertical levels and three constant fields, which amounts to $114$ channels, are stacked to create the input signal.
     \item FourCastNet \cite{fourcasenet}. FourCastNet is a Fourier-based neural network with a vision transformer (ViT) backbone \cite{vit}. It models $20$ variables at five vertical levels and a resolution of $0.25^\circ$ to generate forecasts of variables, including U10, V10, T2M, Z500, T850, and TP.
 \end{enumerate}

\subsection{Evaluation Metrics}
Following \cite{weatherbench}, we use latitude-weighted RMSE (root-mean-square error) 
to evaluate model performance. 
The latitude-weighted RMSE is calculated as follows.

\begin{equation}
    RMSE = \frac{1}{N_{forecasts}} \sum_i^{N_{forecasts}} \sqrt{\frac{1}{N_{lat} N_{lon}} \sum_j^{N_{lat}} \sum_k^{N_{lon}} {L(j) (f_{i,j,k} - t_{i,j,k})}^2}
\end{equation}
where $f$ and $t$ denotes the model forecast and ERA5 truth, respectively. $L(j)$ denotes the weighting coefficient of the $j$th latitude index, which assigns larger weights to lower latitudes  and smaller weights to higher latitudes
as follows. 
\begin{equation}
    L(j) = \frac{\cos{lat(j)}}{\frac{1}{N_{lat}} \sum_j^{N_{lat}} \cos{lat(j)}}.
\end{equation}
We also compute ACC (anomaly correlation coefficient) metric for additional evaluation, which can be found in the support information.
In addition, we adopt the continuous ranked probability score (CRPS) to measure the calibration and sharpness of the ensemble forecasts. CRPS is computed as
\begin{equation}
    CRPS = \int_{-\infty}^\infty (P_{fcst}(x) - P_{obs}(x))^2 dx
\end{equation}
where $P_{fcst}$ and $P_{obs}$ denote the CDF of the forecast distribution and the truth distribution, respectively. For a deterministic forecast, the CRPS reduces to the mean absolute error. The value range of CRPS is greater than or equal to $0$, and the lower the better. In addition, we also plot the rank histograms to measure the ensemble reliability, which can be found in the support information.



\begin{table}[t]
\centering
\caption{Ablation Study for the Designed Components of SwinRNN on WeatherBench Dataset.}
\begin{tabular}{c c c c c c c}
\toprule
& & & \multicolumn{4}{c}{RMSE (5 days)} \\
\cmidrule{4-7}
Residual Add & Multi-sclae & Swin Attention & Z500 ($m^2s^{-2}$) & T850 (K) & T2M (K) & TP (mm) \\
\midrule
\checkmark & \checkmark & \checkmark & $428$ & $2.20$ & $1.75$ & $2.17$  \\
\checkmark & & \checkmark & $431$ & $2.22$ & $1.80$ & $2.21$  \\
& \checkmark & \checkmark & $450$ & $2.29$ & $1.88$ & $2.21$ \\
\checkmark & \checkmark & & $489$ & $2.41$ & $1.88$ & $2.18$ \\
\midrule
\multicolumn{7}{l}{\begin{tabular}{l}
\emph{Note.} Latitude-weighted RMSE scores for 5 day forecasts of Z500, T850, T2M, and TP are reported. \\ The lower the better.
\end{tabular}} \\
\end{tabular}
\label{table-ablation-swinrnn}
\end{table}

\subsection{Ablation Study} \label{sec-ablation-study}
In this subsection, we first conduct ablation study on SwinRNN to investigate the effectiveness of the specially designed components (i.e. the multi-scale decoder, residual prediction, and Swin attention based memory update mechanism) by removing each component from the model each time and comparing the forecast accuracy. All the ablation experiments use the same training setting with $32$ batch size and $30$ epoch.
Then, we conduct several comparison experiments to study the effect of the proposed perturbation module in SwinVRNN, 
including the effect of learning the covariance, and the influence of different numbers of ensemble members.
Finally, we perform comparison experiments on different ensemble forecasting approaches, i.e. fixed distribution perturbation, learned distribution perturbation, model parameters perturbation, and multi-model ensemble.
We maintain the same setting for all the SwinVRNN ablations and ensemble forecasting experiments with $16$ batch size and $30$ epoch, due to the memory limit.


\subsubsection{SwinRNN}

We first evaluate the effect of the multi-scale decoder. We prune the lower three decoders with $\frac{1}{2}$, $\frac{1}{4}$, and $\frac{1}{8}$ scales (as shown in Figure \ref{fig-swinvrnn} (a)) and retain the top decoder with the original scale unchanged.
As shown in Table \ref{table-ablation-swinrnn} Row $1$ and Row $2$, removing multi-scale decoders from the baseline results in slight increase of $5$ day RMSE on all the four variables. 
It shows the positive effect of forecasting atmospheric fields at different resolutions.
We argue that multi-scale forecasting would achieve greater improvements when predicting higher resolution inputs, such as $1$ or $0.25$ degree.

Several previous methods \cite{fourcasenet, gnn-paper} adopt the residual prediction strategy by default. In this paper, we study the influence of residual prediction by directly forecasting variable states themselves instead of their $6$ hour changes.
As shown in Table \ref{table-ablation-swinrnn} Row $1$ and Row $3$, the performance degrades significantly without adding residual operation. 
The $5$ day RMSE of Z500, T850, T2M, and TP increase by $5.2\%$, $4.1\%$, $7.4\%$, and $1.8\%$, respectively. This shows that it is much easier for the model to learn the variables' change with the residual connection, 
since it provides an initial guess of the next states with the previous ones.

Finally, to verify the effectiveness of the proposed Swin attention based memory update mechanism, we replace it with ConvGRU \cite{convgru} architecture while keeping other modules unchanged. 
Row $1$ and Row $4$ in Table \ref{table-ablation-swinrnn} show the comparison results. The performance of all the four variables declines considerably, which justifies the superiority of leveraging the self-attention operation to model the relation between the historical hidden representation and the current inputs to the gate mechanism of ConvGRU.




\subsubsection{SwinVRNN} \label{sec-ablation-swinvrnn}

\begin{table}[t]
 \caption{Effect of Learning the Covariance Matrix.}
 \centering
 \begin{tabular}{l c c c c c}
 \toprule
    & & \multicolumn{4}{c}{RMSE (5 days/14 days)} \\
    \cmidrule{3-6}
     Model & Covariance & Z500 ($m^2s^{-2}$) & T850 (K) & T2M (K) & TP (mm) \\
 \midrule
  SwinRNN & & 431/972 & 2.20/4.25 & 1.75/3.28 & 2.17/2.40 \\
  \midrule
  & N & 417/816 & 2.14/3.56 & 1.72/2.90 & 2.18/2.31 \\
  SwinVRNN & Y & $409$/$803$ & $2.10$/$3.51$ & $1.68$/$2.73$ & $2.16$/$2.31$ \\
 \midrule
 \multicolumn{6}{l}{\begin{tabular}{l}
      \emph{Note.} Latitude-weighted RMSE scores for 5 day and 14 day forecasts of Z500, T850, \\ T2M, and TP are reported. The lower the better.
 \end{tabular}}
 \end{tabular}
 \label{table-swinvrnn-ablation}
 \end{table}

\begin{figure}[t]
\centering
\includegraphics[width=13cm]{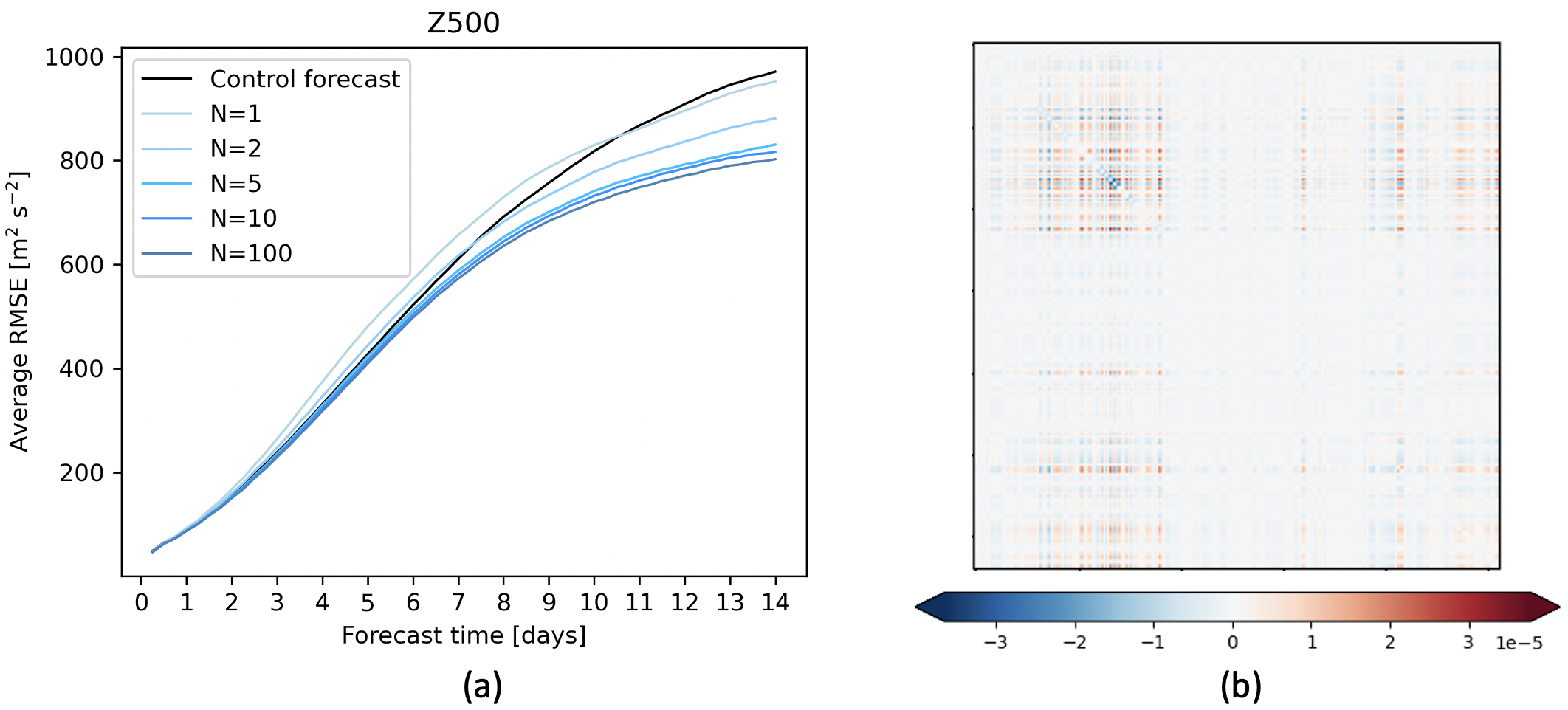}
\caption{(a) The ensemble forecasting performance of SwinVRNN over forecast time with different ensemble members in terms of Z500 field. Each blue line represents the average RMSE scores of multiple samplings. The black line represents the control forecast of the deterministic model SwinRNN. (b) Example of a learned covariance matrix with shape $HW \times HW$, where $H$ and $W$ denote the spatial size of the feature map. We set the diagonal elements to zeros for better visualization, since the variances and covariances have different magnitude ($1e0$ and $1e-5$ respectively.)
\label{fig-members}}
\end{figure}

First, we justify the effectiveness of learning the covariance between the random variables in the spatial dimension when modeling the multivariate Gaussian distribution of $z_t$.
Table \ref{table-swinvrnn-ablation} shows the comparison results at 5 day and 14 day forecasts. All the results of SwinVRNN model are the ensemble forecast with $100$ members, and the result of SwinRNN is the control forecast without perturbation.
As shown in Table \ref{table-swinvrnn-ablation} Row 2 and Row 3, SwinVRNN ignoring the covariance already outperforms the SwinRNN baseline in almost all the variables, and learning the covariances further improves the performance significantly. 
The experimental results comply with the spatial autocorrelation assumption described by the first law of geography (Section \ref{sec-perturbation-module}). Random variables in different spatial locations are not ideally independent, thus considering their correlation when sampling the noise for a specific pixel location is necessary.
Figure \ref{fig-members} (b) shows an example of a learned covariance matrix.
The diagonal of the covariance matrix represents the variances of all latent variables in the spatial dimension, which have values around $1$, 
while the covariances of different variables are in magnitude of $1e-5$.
Thus, to better visualize the pattern of covariances, we set the diagonal values to zeros.
Each line of the matrix represents the covariance between a pixel and other pixels of the feature map. 
As can be seen, some pixels are activated with positive correlation (red) or negative correlation (blue), which correspond to the locations with close distances to the current pixel, while most other locations have covariance close to zero, which correspond to most distant pixels.
In addition, the covariance matrices for different channels and different time steps have different patterns, in which a pixel may have high correlations with the pixels in a large vicinity or only a small neighborhood.
However, in general, the covariance matrix is a sparse matrix with the covariance between most distant pixels being close to zero.

In addition, we investigate the ensemble performance of different numbers of ensemble members. 
As shown in Figure \ref{fig-members} (a), the average RMSE gradually declines with the increase of the ensemble member.
Especially, the improvement is significant when the ensemble number is less than $10$ members. Moreover, the improvement of ensemble forecasting is more apparent on $14$ day forecast than that on $5$ day forecast. For example, when the ensemble number increases from $1$ to $2$, the average RMSE scores of $5$ day and $14$ forecast degrade from $480/951$ to $444/881$ for Z500. Furthermore, the ensemble mean of $2$ members approaches and surpasses the performance of control forecast of the deterministic model SwinRNN within $5$ days and after $6$ days, respectively. And only one member of stochastic SwinVRNN exceeds SwinRNN after $10$ days.
There are three conclusions can be drawn from the phenomenon: 
1) Ensemble forecasting using SwinVRNN is able to achieve reasonable ensemble spread, which accounts for the significant improvement on ensemble mean with a small number of members. This is further discussed in Section \ref{sec-verification}.
2) The ensemble performance gradually improves with the increase of forecast lead time due to the increasing uncertainty of the future.
3) SwinVRNN considering stochasticity on each step of future forecasting is more reasonable than the deterministic SwinRNN model.

\subsubsection{Ensemble Forecasting} \label{sec-ablation-ensemble}

In this subsection, we introduce the four different categories of data-driven based ensemble forecasting methods as mentioned in the introduction section and compare their ensemble performance.
1) Fixed distribution perturbation. It is achieved by perturbing the input state variables with Gaussian random noise drawn from a fixed standard normal distribution. For example, FourCastNet \cite{fourcasenet} utilizes this method to perturb the initial conditions, simulating the way operational NWP models generate perturbed initial conditions.
In this paper, we take advantage of this method to perturb the input $x_t$ in each time step of future states forecasting.
A fixed distribution perturbation is implemented as $x_t = x_t + \sigma \xi$, where $x_t$ is the last frame of the history sequence in the first time step, and the output of the previous time step forecasting in the following time steps. 
This accounts for the error introduced by imperfect initial conditions and inaccurate future forecasts, respectively.
$\xi \sim \mathcal{N}(\mathbf{0}, \mathbf{I})$ is a random normal noise with the same shape as $x_t$, and $\sigma$ is the scaling factor for controlling the intensity of the perturbation. We set $\sigma$ to $0.02$ in this paper.
(2) Learned distribution perturbation. The perturbation noises applied to different variables and different spatial locations are drawn from a fixed Gaussian distribution in the above perturbation method, which is not optimal and reasonable. In addition, the autocorrelation of random variables at different spatial locations are not taken into account. 
Therefore, we propose a perturbation module to learn the distribution from the data, specifically the mean and covariance matrix of the multivariate normal distribution. 
With such a distribution learning scheme, the noise intensity for different channels and different locations is varied dynamically conditioned on the data, and the correlations of latent variables in the spatial dimension are characterized by the covariance matrix. 
The proposed SwinVRNN adopts this perturbation fashion.
(3) Model parameters perturbation. This type of approach is to perturb the model parameters, which shares some spirits with the perturbed parameter schemes in modern NWP models \cite{mccabe2016representing}. Here, we leverage the Monte Carlo dropout (MC dropout) \cite{dropout} to achieve a simple model parameters perturbation for data-driven models. A dropout layer is applied to the input or the hidden feature to stochastically remove some neuron connections. Note that the dropout operation in MC estimate works for both training and inference phases, which differs from the way it is usually used in other deep learning models in which dropout is turned off during inference.
(4) Multi model ensemble. Another readily achievable ensemble forecasting method is multi-model ensemble, which is usually used in improving prediction accuracy at test stage, since the average result of multiple models is usually superior to that of a single model. Multiple models can be models with totally different architectures, but in this paper, we obtain different models by simply tuning the architecture of SwinVRNN, such as the depth of the Swin decoder.

\begin{table}[t]
 \caption{Comparison of Different Ensemble Forecasting Methods on WeatherBench Dataset.}
 \resizebox{\textwidth}{!}{
 \centering
 \begin{tabular}{l c c c c c | c c c c}
 \toprule
    & & \multicolumn{4}{c}{RMSE (5 days/14 days)} & \multicolumn{4}{c}{CRPS (5 days/14 days)} \\
    \cmidrule{3-10}
    Backbone & Perturbation Method & Z500 ($m^2s^{-2}$) & T850 (K) & T2M (K) & TP (mm) & Z500 ($m^2s^{-2}$) & T850 (K) & T2M (K) & TP (mm) \\
 \midrule
  SwinRNN  & - & 431/972 & 2.20/4.25 & 1.75/3.28 & 2.17/2.40 & 339/764 & 1.76/3.44 & 1.40/2.63 & 0.52/0.66 \\
  SwinRNN & fixed distribution & 431/952 & 2.20/4.13 & 1.75/3.25 & 2.17/2.36 & 303/646 & 1.66/2.94 & 1.27/2.28 & 0.50/0.58 \\
  SwinRNN & MC dropout & 426/909 & 2.18/3.94 & 1.73/3.08 & 2.18/2.32 & 306/597 & 1.59/2.72 & 1.25/2.25 & 0.49/0.52 \\
  SwinVRNN & learned distribution & 409/803 & 2.10/3.51 & 1.68/2.73 & 2.16/2.31 & 242/483 & 1.31/2.18 & 1.05/1.67 & 0.45/0.49 \\
  SwinVRNN & multi-model & \textbf{401/795} & \textbf{2.07/3.47} & \textbf{1.64/2.70} & \textbf{2.16/2.30} & \textbf{232/473} & \textbf{1.26/2.14} & \textbf{0.99/1.63} & \textbf{0.44/0.48} \\
 \midrule
 \multicolumn{10}{l}{\begin{tabular}{l}
      \emph{Note.} Latitude-weighted RMSE and CRPS for 5 day and 14 day forecasts of Z500, T850, T2M, and TP are reported. The lower the better.
 \end{tabular}}
 \end{tabular}
 \label{table-ensemble-forecast}
 }
 \end{table}
 
\begin{figure}[t]
\centering
\includegraphics[width=13cm]{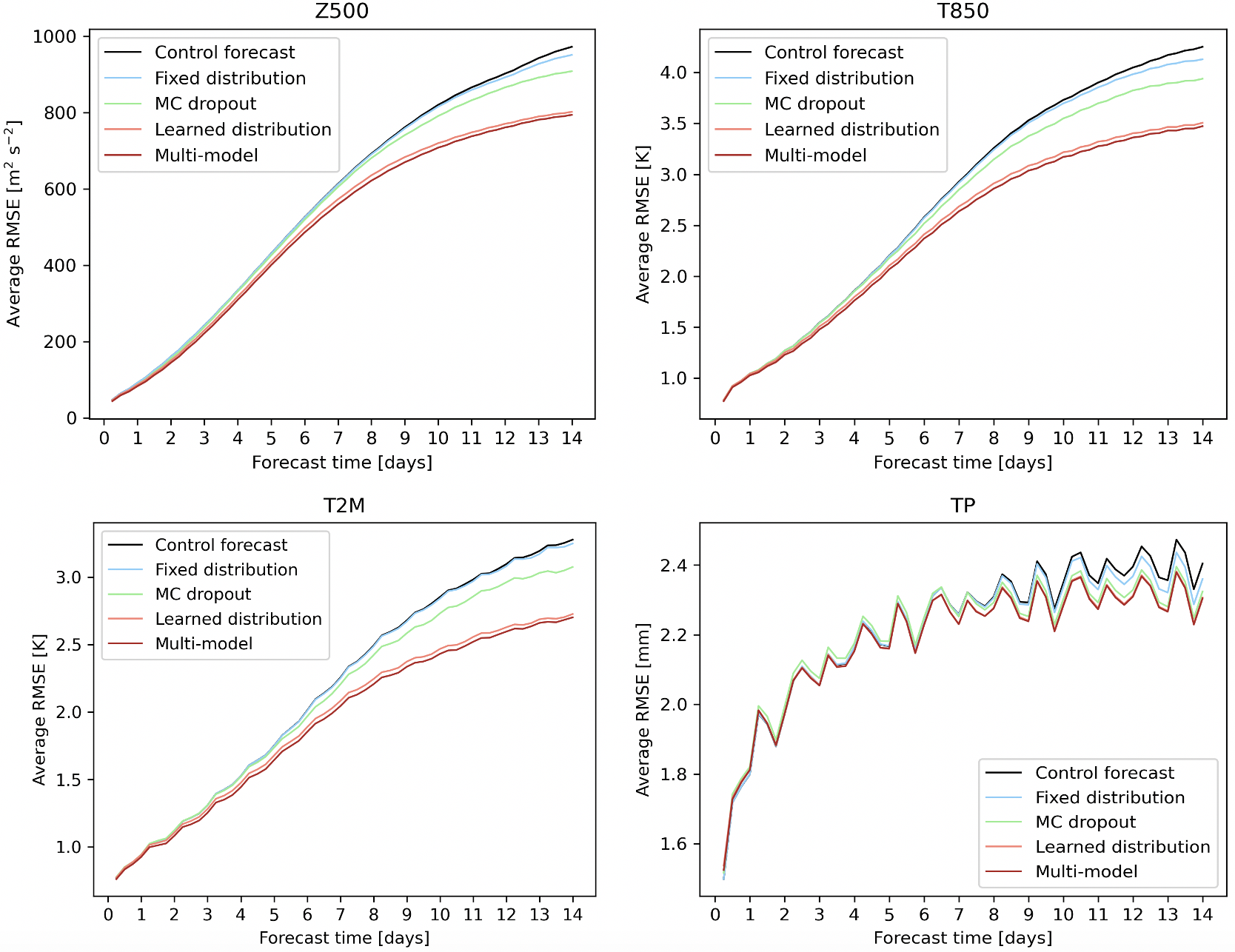}
\caption{Comparison of four different perturbation methods, i.e. fixed distribution perturbation, MC dropout, learned distribution perturbation, and multi-model ensemble. MC dropout is a model parameter perturbation method. Learned distribution perturbation refers to the proposed SwinVRNN model. Control forecast is generated by the deterministic SwinRNN model without perturbation. \label{fig-ensemble-forecast}}
\end{figure}

Table \ref{table-ensemble-forecast} shows the $5$ day and $14$ day results of the above four ensemble forecasting approaches. All the models are trained for $5$ day forecasting, and the $14$ day results are reported to show the models' generalization of extrapolating to longer lead time. Furthermore, all the ensemble forecasting results are averages of $100$ members, except that the multi-model ensemble method aggregates $300$ members from three SwinVRNN models where each model generates $100$ members. The SwinRNN without perturbation is compared as a baseline for control forecast.

We first evaluate the forecast accuracy (RMSE) of the simplest fixed distribution perturbation method. As shown in Table \ref{table-ensemble-forecast} Row $1$ and Row $2$, the RMSE scores of $5$ day forecast for all the four variables have no change, but the $14$ day forecast performance gets some small improvements. Figure \ref{fig-ensemble-forecast} also shows the trend that fixed distribution perturbation starts to improve the control forecast baseline from very late lead time, such as around 10 days for Z500, and 12 days for T2M.
We then study the effect of MC dropout. As can be seen from Table \ref{table-ensemble-forecast} Row $1$ and Row $3$, both of the $5$ day and $14$ day forecast performance gets non-trivial improvements, especially the $14$ day results, except the $5$ day forecast of TP. 
Figure \ref{fig-ensemble-forecast} shows the MC dropout method exceeds the baseline earlier than the fixed distribution method, which demonstrates the effectiveness of perturbing the model parameters using dropout operation. 
The MC dropout method is simple to implement and able to improve the forecasting performance within the training period and extrapolation times.
Table \ref{table-ensemble-forecast} Row $4$ shows the results of our proposed learned distribution method by perturbing the hidden features. The performance of all the evaluated four variables achieves remarkable improvements for both $5$ days and $14$ days forecasts. Specifically, the RMSE scores decrease from $431/972$ to $409/803$ for Z500, $2.20/4.25$ to $2.10/3.51$ for T850, $1.75/3.28$ to $1.68/2.73$ for T2M, and $2.17/2.40$ to $2.16/2.31$ for TP. Figure \ref{fig-ensemble-forecast} also shows a clear margin between our method with the control forecast and the above two mentioned ensemble methods.
The ensemble results of our SwinVRNN model significantly outperform those of the control baseline even within the $5$ days. This can be attributed to its reasonable spread which will be discussed in the next subsection.
It justifies the effectiveness and superiority of learning the noise distribution from the data using the proposed variational perturbation module. 
The last row in Table \ref{table-ensemble-forecast} shows the three-model ensemble results of SwinVRNN.
By aggregating the forecasting members from different SwinVRNN models, the performance can be further boosted for both $5$ days and $14$ days forecasts. For example, the RMSE scores degrade from $409/803$ to $401/795$ for Z500, $2.10/3.51$ to $2.07/3.47$ for T850. Note that the RMSE curve of TP fluctuates violently and the trend is a little different from the other three variables, which may be due to the discontinuity of precipitation along the time dimension and the sparse property in the spatial dimension. 

\subsubsection{Ensemble Verification}
\label{sec-verification}

\begin{figure}[t]
\centering
\includegraphics[width=16cm]{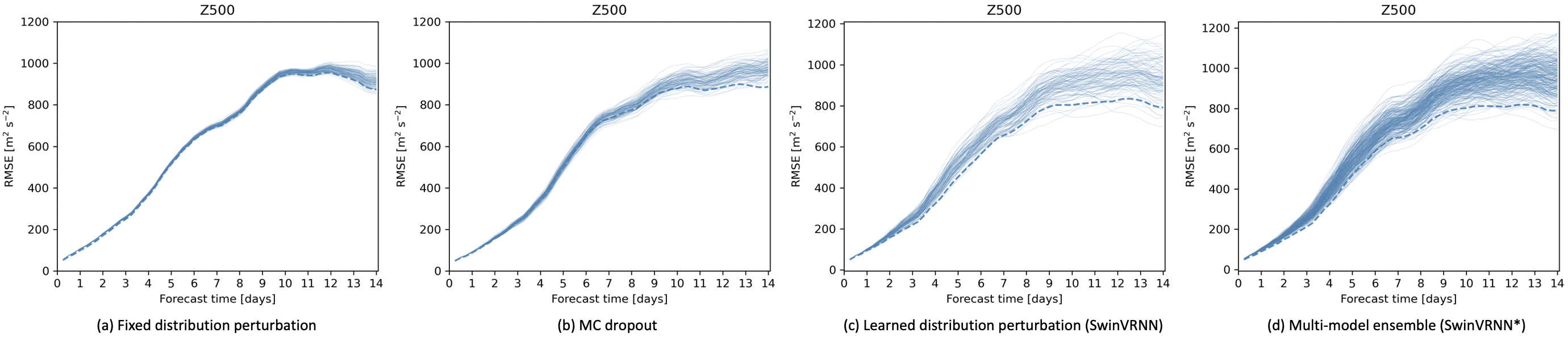}
\caption{RMSE spread of the four perturbation methods, i.e. fixed distribution perturbation, MC dropout, learned distribution perturbation, and multi-model ensemble. Each thin line represents an ensemble member. $100$ members are drawn for the first three single-model perturbation methods and $300$ members are drawn for the multi-model ensemble method. The dashed line represents the ensemble mean. \label{fig-rmse-spread-of-four-methods}}
\end{figure}

\begin{figure}[t]
\centering
\includegraphics[width=16cm]{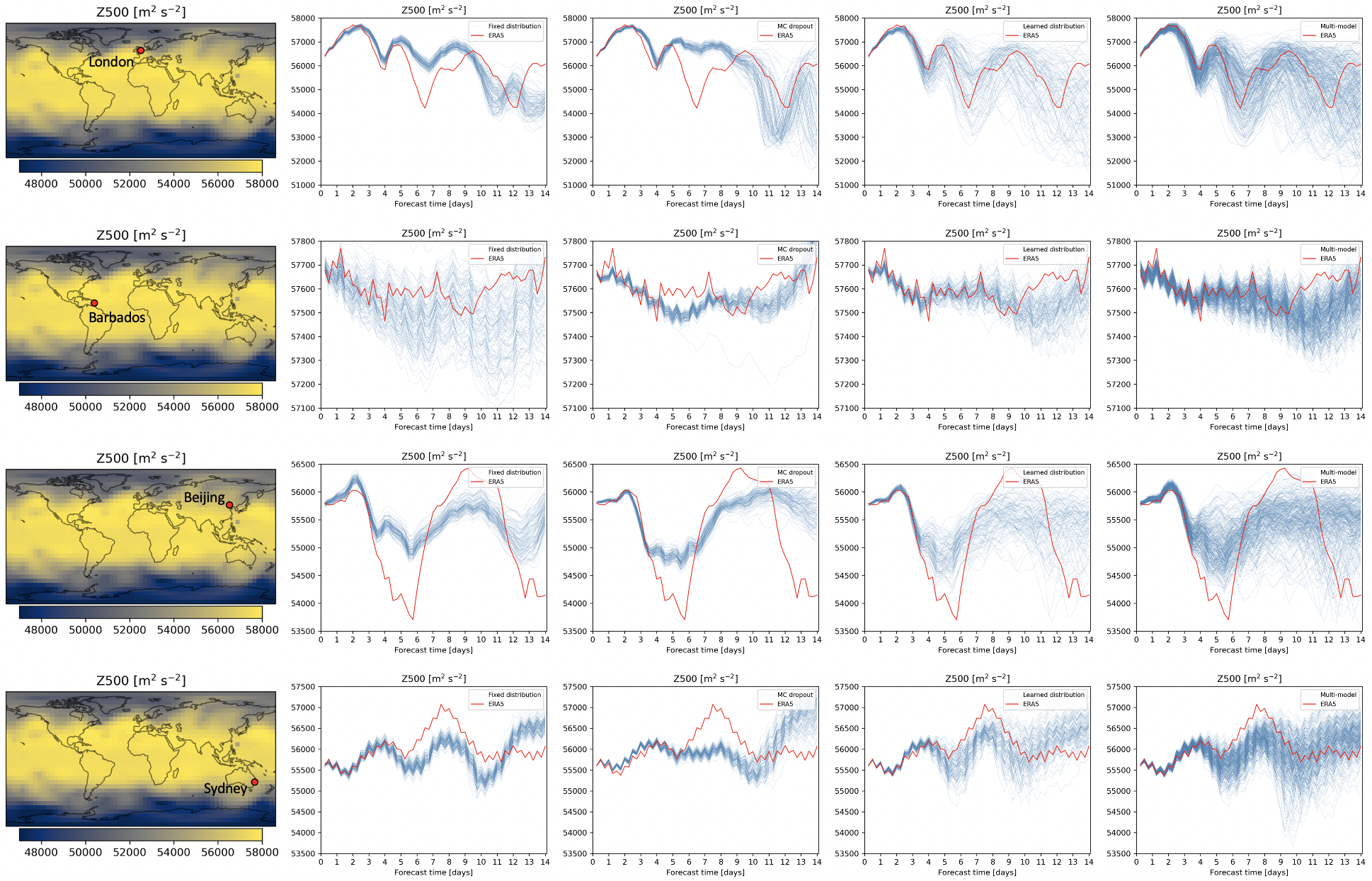}
\caption{Ensemble spread for Z500 prediction over London, Barbados, Beijing and Sydney. The First column shows the studied locations on Z500 field. Column $2$ to Column $5$ show the $100$ predictions of the fixed distribution perturbation method, MC dropout, learned distribution perturbation method, and multi-model ensemble, respectively. The red curve denotes the ERA5 truth.
\label{fig-ensemble-spread-of-four-citys-on-z500}}
\end{figure}

First, we visualize the RMSE spread of the four perturbation methods. Figure \ref{fig-rmse-spread-of-four-methods} shows the RMSE spread with each thin line representing an ensemble member and the dashed line representing the ensemble mean. Several phenomena can be observed from the figure. 
1) In terms of the RMSE spread, multi-model ensemble (represented by SwinVRNN$^*$) is greater than the learned distribution perturbation (represented by SwinVRNN), SwinVRNN is greater than MC dropout, and MC dropout is greater than the fixed distribution perturbation.
2) The ensemble mean is greatly influenced by the RMSE spread.
As shown in Figure \ref{fig-rmse-spread-of-four-methods} (a), the RMSE spread of the fixed distribution method is extremely limited within $10$ days and gradually expands afterward. Accordingly, the RMSE of ensemble mean is not better than any ensemble member within $10$ days,
but afterward, the RMSE of ensemble mean gradually achieves improvements and is lower than most ensemble members (its curve is located at the bottom of the cluster), with the increase of the spread.
This phenomenon is also in accordance with the Z500 results of the fixed distribution method in Figure \ref{fig-ensemble-forecast}. 
As shown in Figure \ref{fig-ensemble-forecast}, fixed distribution has no improvement within $10$ days compared to the control forecast, but gradually gains promotion afterwards. 
Figure \ref{fig-rmse-spread-of-four-methods} (b) shows that the MC dropout has greater RMSE spread and better ensemble mean than the fixed distribution method, which is also consistent with Figure \ref{fig-ensemble-forecast}. 
As can be seen from Figure \ref{fig-rmse-spread-of-four-methods} (c), the proposed SwinVRNN method has much greater RMSE spread in all lead times up to $14$ days, and the RMSE of the ensemble mean is significantly superior to that of most individual members (its RMSE curve is located much lower than most members). 
The great spread within $5$ days explains the remarkable performance gain of SwinVRNN even within the training window of forecast time. 
Figure \ref{fig-rmse-spread-of-four-methods} (d) shows that SwinVRNN$^*$ with $300$ members has wider spread and greater ensemble mean compared with SwinVRNN.

In addition, we visualize the ensemble spread of the four methods on Z500 prediction in Figure \ref{fig-ensemble-spread-of-four-citys-on-z500}. Four cities (London, Barbados, Beijing, and Sydney) are chosen as study cases. For each city, we plot the prediction curves of the four methods as well as the ERA5 truth.
Each method's $100$ predictions are plotted with blue curves, and the truths are plotted with red curves.
As shown in Figure \ref{fig-ensemble-spread-of-four-citys-on-z500} Column $2$, two extremes are displayed.
For London, Beijing, and Sydney, the ensemble spread of the fixed distribution method is narrow, and can not cover the truth. 
Moreover, the curve overall tendency of predictions does not agree with the truth. For example, the dramatic drop of Z500 in Beijing at the lead time around $6$ days is not reflected in all the $100$ predictions.
However, for Barbados, the prediction spread almost unreasonably explodes. Although the spread covers the truth, the prediction tendency is inconsistent. 
We argue that the extremely limited or extremely large ensemble spread of the fixed distribution method is caused by the unadjustable perturbation intensity. 
For cities like London, Beijing and Sydney, Z500 field fluctuates severely with time, but the preset noise intensity is relatively small for regions like these cities, thus leading to limited ensemble spread.
However, for cities like Barbados, the change of Z500 field is much smoother along time, but the preset noise intensity is too strong, which thus results in exploded spread.
As for MC dropout, as shown in Figure \ref{fig-ensemble-spread-of-four-citys-on-z500} Column $3$, the ensemble spread for all the studied cities is small and can hardly cover the truth, especially in the regions with rapid changes.
This also demonstrates the unsatisfied ensemble performance of MC dropout. 
Figure \ref{fig-ensemble-spread-of-four-citys-on-z500} Column $4$ shows the results of our learned distribution method. For all the studied cities, our model can generate large and reasonable ensemble spread, which can cover the truth very well in most situations.
For example, for London, the predictions spread around the truth in all lead times, and the tendency perfectly agrees with the truth.
Nevertheless, in regions with severe fluctuation, such as Beijing at lead time around $6$ days, the predictions do not reach the bottom of the curve, but are much better than previous two methods. 
The last column of Figure \ref{fig-ensemble-spread-of-four-citys-on-z500} shows the results of the multi-model ensemble method. It can be seen that the ensemble spread further expands and has better coverage of the truth compared with the single model counterpart.

The forecast calibration and sharpness evaluated by CRPS of the four methods are shown in the right part of Table \ref{table-ensemble-forecast}. $5$ day and $14$ day forecasts are reported. Table \ref{table-ensemble-forecast} Row $1$ shows the results of the unperturbed control forecast baseline, whose CRPS is equivalent to MAE. 
It scores worse due to none probabilistic information.
Row $2$ to Row $4$ in Table \ref{table-ensemble-forecast} show the results of three single model perturbation methods. Among them, the proposed learned distribution outperforms the other two methods significantly in all the four variables. Again, multi-model ensemble in the last row further achieves non-trivial improvements.

\subsection{Comparison with SOTA} \label{sec-exp-sota}

\begin{table}[t]
 \caption{Comparison with Baselines on WeatherBench Dataset.}
 \centering
 \begin{tabular}{l c c c c}
 \toprule
    & \multicolumn{4}{c}{RMSE (3 days/ 5 days/ 14 days)} \\
    \cmidrule{2-5}
    Method & Z500 ($m^2s^{-2}$) & T850 (K) & T2M (K) & TP (mm) \\
 \midrule
 Weekly climatology & $816$ & $3.50$ & $3.19$ & $2.32$ \\
 \midrule
 T42 & $489$/$743$/$-$ & $3.09$/$3.83$/$-$ & $3.21$/$3.69$/$-$ \\
 T63 & $268$/$463$/$-$ & $1.85$/$2.52$/$-$ & $2.04$/$2.44$/$-$ \\
 IFS & \textbf{154/334}/$-$ & \textbf{1.36}/$2.03$/$-$ & $1.35$/$1.77$/$-$ & $2.36$/$2.59$/$-$ \\
 \midrule
 Na\"ive CNN & $626$/$757$/$-$ & $2.87$/$3.37$/$-$ \\
 Cubed UNet & $373$/$611$/$-$ & $1.98$/$2.87$/$-$ \\
 ResNet (pretrained) & $284$/$499$/$-$ & $1.72$/$2.41$/$-$ & $1.48$/$1.92$/$-$ & $2.23$/$2.33$/$-$ \\
 FourCastNet & $240$/$480$/$-$ & $1.50$/$2.50$/$-$ & $1.50$/$2.00$/$-$ & $2.20$/$2.50$/$-$ \\
 \midrule
 SwinRNN & $207$/$392$/$882$ & $1.39$/$2.05$/$3.90$ & \textbf{1.18}/$1.63$/$2.98$ & \textbf{2.01/2.14}/$2.34$ \\
 SwinVRNN & $219$/$397$/$788$ & $1.47$/$2.06$/$3.45$ & $1.25$/$1.66$/$2.69$ & $2.06$/$2.17$/\textbf{2.31} \\
 SwinVRNN$^*$ & $211$/$388$/$783$ & $1.43$/\textbf{2.02/3.43} & $1.21$/\textbf{1.61/2.67} & $2.05$/$2.16$/\textbf{2.31} \\
 \midrule
 \multicolumn{5}{l}{\begin{tabular}{l}
      \emph{Note.} Latitude-weighted RMSE scores for 3 day, 5 day forecasts of Z500, T850, T2M, and TP \\ are reported. The lower the better. SwinVRNN$^*$ denotes multi-model ensemble of SwinVRNN.
 \end{tabular}}
 \end{tabular}
 \label{table-comp-sota}
 \end{table}
 
\begin{figure}[t]
\centering
\includegraphics[width=14cm]{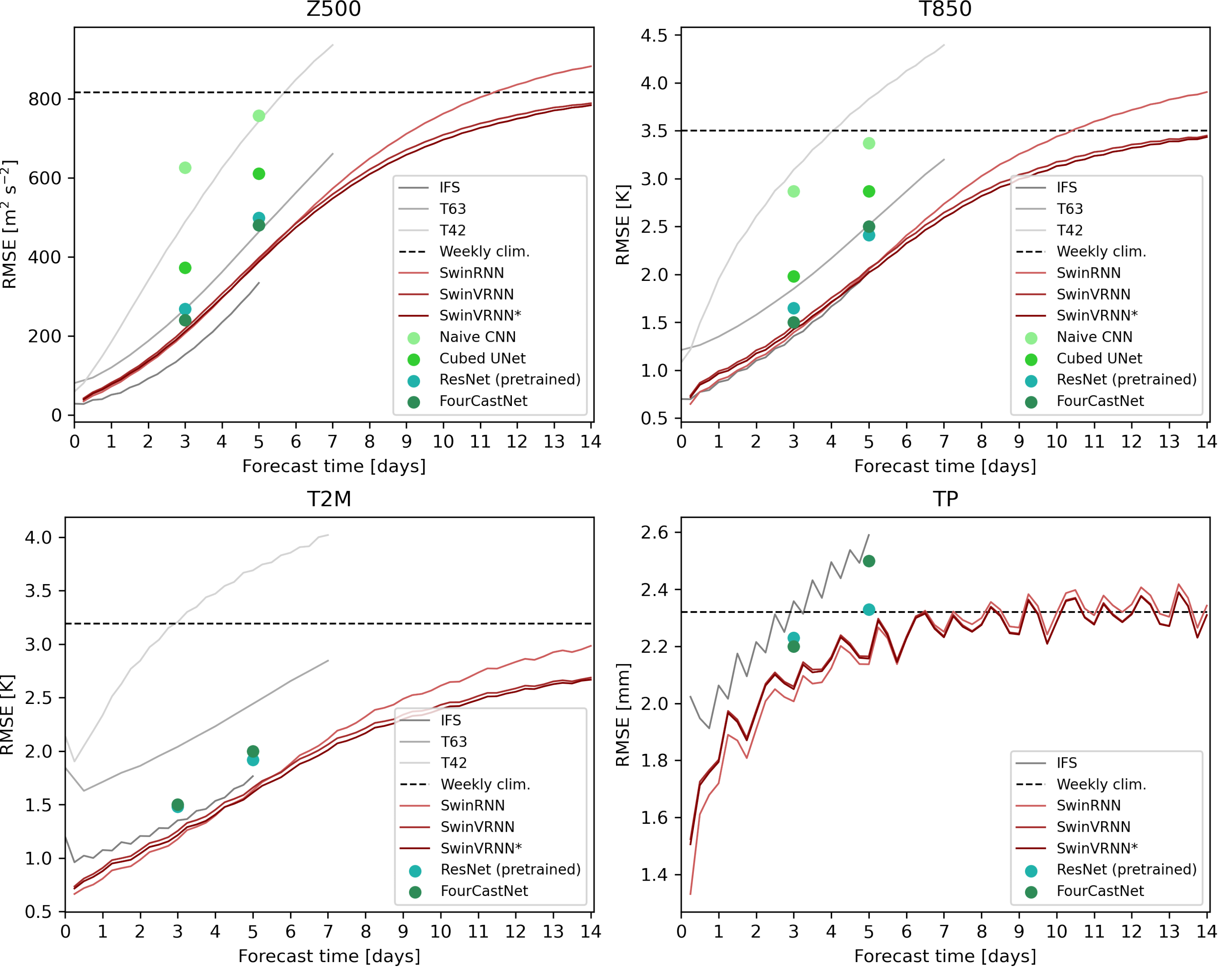}
\caption{Latitude-weighted RMSE of Z500, T850, T2M, and TP for the proposed SwinRNN, SwinVRNN, and SwinVRNN$^*$ models and three categories of baselines. Data-driven baselines (Na\"ive CNN, Cubed UNet, ResNet (pretrained), and FourCastNet) are denoted with dots in different colors, showing the RMSE score of $3$ day and $5$ day forecasts. Weekly climatology is denoted with a dashed horizontal line. NWP baselines (T42, T63, and NWP) are denoted by gray lines, and the proposed models (SwinRNN, SwinVRNN, and SwinVRNN$^*$) are denoted by red lines.
\label{fig-sota}}
\end{figure}

\begin{figure}[t]
\centering
\includegraphics[width=16cm]{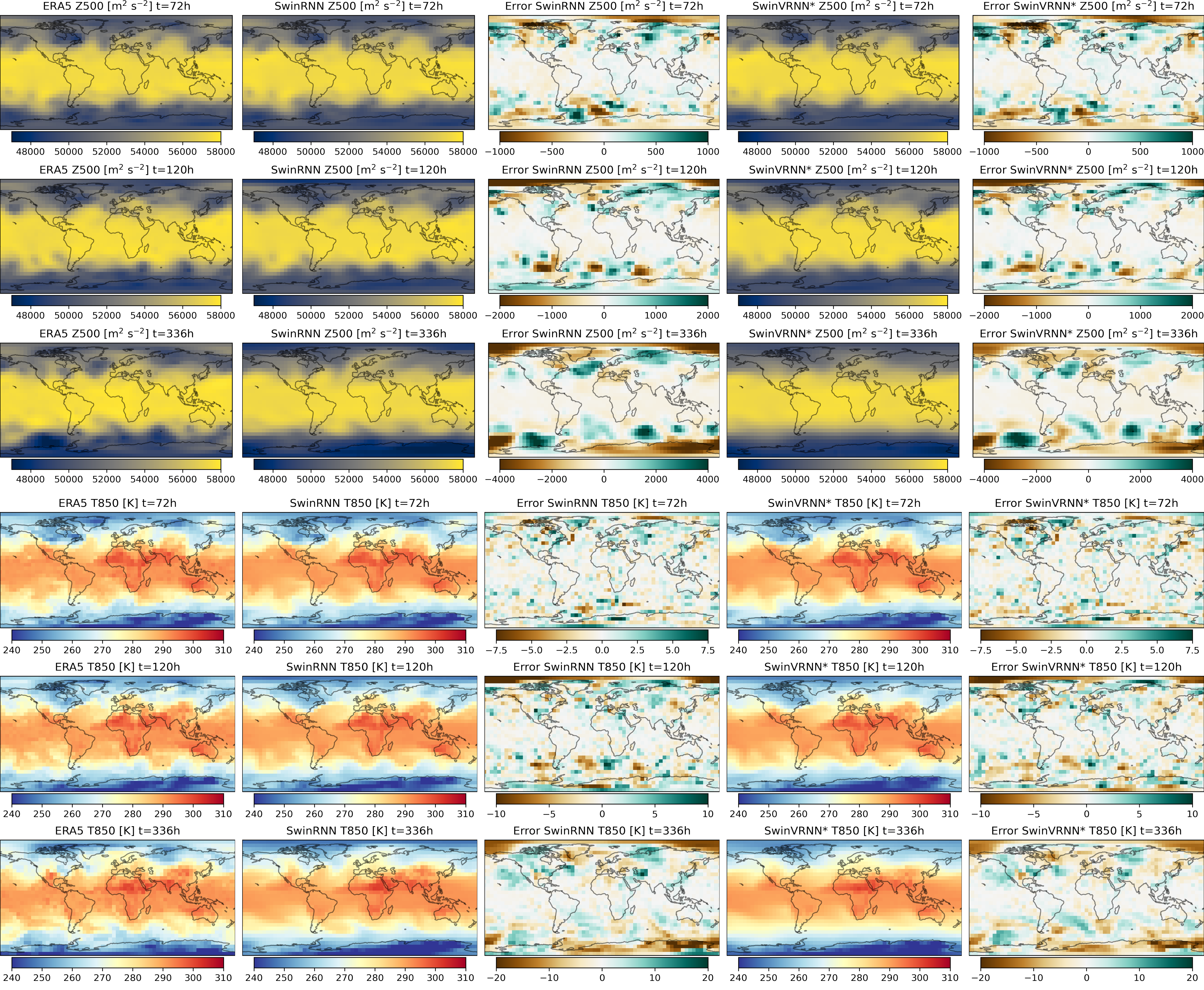}
\caption{Qualitative comparison among ERA5 truth, SwinRNN, and SwinVRNN$^*$ on example fields of Z500 and T850 initialized at 11 UTC 3 August 2018. The top three and bottom three rows show the fields at 72h, 120h, and 336h for Z500 and T850, respectively. The first column shows the ERA5 truth. Column 2 and Column 4 show the control forecast of deterministic SwinRNN and ensemble forecast of SwinVRNN$^*$, respectively. In addition, we display the difference between the forecast results of SwinRNN and SwinVRNN$^*$ with the corresponding ERA5 truth at Column 3 and Column 5 to facilitate the analysis of their forecasting quality.
\label{fig-example}}
\end{figure}

We compare the proposed SwinRNN, SwinVRNN, and SwinVRNN$^*$ with climatology baseline, NWP baselines, and data-driven (deep learning weather prediction) baselines on WeatherBench dataset. SwinVRNN$^*$ denotes the three-model ensemble of SwinVRNN with decoder depth $4$, $5$, and $6$ trained with the same schedule. For comparison with state-of-the-art methods, we train our models for a longer schedule using the two-phase strategy illustrated in Section \ref{sec-training}.

Table \ref{table-comp-sota} shows the comparison results between our models with three categories of baselines at forecast lead time of $3$ days, $5$ days and $14$ days.
$14$-day forecast results for baseline models are unavailable, thus remaining vacant. 
Figure \ref{fig-sota} shows the curve of RMSE over forecast lead time. In the figure, data-driven baselines are denoted with dots in different colors, showing the RMSE score of 3-day and 5-day forecasts. Weekly climatology is denoted with a dashed horizontal line. NWP models and our models are represented with curves in different colors. 
WeatherBench provides 7-day forecasts of Z500, T850, and T2M fields for T42 and T63 models, and 5-day forecasts of all the four studied fields (including TP) for IFS. We display the forecasting performance until 14 days for our models, in which the first 5 days are the training period, and next 9 days are the extrapolation lead time.

As shown in Table \ref{table-comp-sota} and Figure \ref{fig-sota}, comparing with data-driven baselines, all of our models achieve superior performance on the four evaluated atmospheric fields (Z500, T850, T2M, and TP). 
The proposed SwinRNN outperforms FourCastNet by a clear margin, especially on 5-day forecast ($392$ vs. $480$ for Z500, $2.05$ vs. $2.50$ for T850, $1.63$ vs. $2.00$ for T2M, and $2.14$ vs. $2.50$ for TP).
The comparison may be unfair due to different input resolution adopted by FourCastNet ($0.25^\circ$) and our models ($5.625^\circ$), and different training window ($12$ hours for FourCastNet and $5$ days for ours).
However, the superior results still demonstrate the effectiveness of our designed Swin transformer based recurrent architecture for weather forecasting. 
When compared with NWP baselines, our models outperform T42 and T63 with a large margin for all variables and all forecast times. 
Regarding the state-of-the-art operational model IFS, our models also achieve breakthrough performance. Specifically, our models outperform IFS on the two surface fields T2M and TP with a clear margin at all the forecasting lead times up to 5 days. For example, the RMSE scores for SwinRNN and IFS at lead time of 3 days and 5 days are $1.18$/$1.63$ versus $1.35$/$1.77$ for T2M, and $2.01$/$2.14$ versus $2.36$/$2.59$ for TP.
As for T850 field, our models achieve comparable results with IFS within 5 days and SwinRNN$^*$ slightly surpasses IFS at 5-day forecast. Specifically, the RMSE scores for SwinRNN$^*$ and IFS at lead time of 3 days and 5 days are $1.43$/$2.02$ versus $1.36$/$2.03$. 
For Z500 field, our models lag behind the IFS, but still set a new state-of-the-art among data-driven models. 
In addition, we compare the performance of control forecast and ensemble forecasts generated by the proposed deterministic SwinRNN model, stochastic SwinVRNN and SwinVRNN$^*$ models. 
As can be seen from Table \ref{table-comp-sota}, the performance of SwinVRNN slightly degrades within the training period (5 days) compared with SwinRNN, but the multi-model version of SwinVRNN$^*$ pulls back some scores. However, in longer lead time after 5 days, as shown in Figure \ref{fig-sota}, SwinVRNN exceeds SwinRNN gradually and creates a clear gap with the increase of forecast times.
For example, the RMSE scores at a lead time of 14 days are $788$ versus $882$ for Z500, $3.45$ versus $3.90$ for T850, and $2.69$ versus $2.98$ for T2M. 
Moreover, SwinVRNN$^*$ further boosts the performance of the single model SwinVRNN in all forecast lead times.
For example, the RMSE scores are $388$ versus $397$ at 5-day forecast for Z500, and $3.43$ versus $3.45$ at 14-day forecast for T850.
It is also worth noting that for Z500 and T850, deterministic SwinRNN performs worse than the weekly climatology at the lead time around $11$ days, but the ensemble mean of SwinVRNN improves the results and achieves superior performance to the weekly climatology even at the lead time of $14$ days.
This further verifies the effectiveness of the proposed learned feature perturbation module.
Overall, our models achieve new state-of-the-art among data-driven based medium-range weather forecasting models. More notably, our models outperform the operational IFS model in all forecast lead times within 5 days on surface fields of T2M and TP and approaches IFS closely and slightly surpasses it at 5 days on T850 field.

Figure \ref{fig-example} shows example fields of Z500 and T850 at 11 UTC 3 August 2018 initialization time. We compare the prediction quality of control forecast and ensemble forecast generated by SwinRNN and SwinVRNN$^*$ at 72h, 120h, and 336h.
To better visualize the forecast quality, we plot the error between ERA5 truth and SwinRNN and SwinVRNN$^*$ in Column $3$ and Column $5$ in Figure \ref{fig-example}, respectively.
As can be seen from the error images, the prediction error is higher in high latitudes than that in middle and low latitudes. In addition, when comparing the errors of SwinRNN and SwinVRNN$^*$, we can observe a reduction of error magnitude, which demonstrates the superior forecast quality of ensemble forecast. However, we can see that SwinVRNN$^*$ predictions are much smoother than SwinRNN due to the ensemble mean procedure.




\section{Discussion}
\subsection{Model Architecture}
In this paper, we propose a novel architecture --- SwinRNN for medium-range weather forecasting. It is a recurrent neural network but different from previous RNN models in the starting time step of the recurrent prediction and the hidden state update mechanism.
We conduct careful ablation experiments to verify the significance of our designs, including multi-scale forecasting, residual prediction, and self-attention based memory update mechanism.
It is expected that our designs and ablation results may inspire follow-up research on developing better model architectures for weather forecasting.
In recent years, different types of DL models are applied on medium-range weather forecasting, including CNN-basd models, transformer-based models, and GNN (graph neural networks)-based models. In terms of forecasting performance of the existing models, transformer-based models \cite{fourcasenet} and GNN-based models \cite{gnn-paper} are superior to CNN-basd models \cite{weatherbench, resnet-paper, cube-paper}. The proposed SwinRNN can be categorized into transformer-based models. Transformer-based models have much more rapid access to long-range even global-range receptive field than CNN models, which is vital for large-scale atmospheric event forecasting. Regarding GNN models, they are more general than CNN models, which makes them more suitable for handling the spherical gemometry of the earth and processing multi-resolution data in one model \cite{gnn-paper}.
Further research would be worthy on the architecture exploration for any of the transformer-based, GNN-based, CNN-based or RNN-based data-driven models.

\subsection{Ensemble Forecasting}
We claim that perturbation can be added to different levels of data-driven based forecasting models, which include but are not limited to input level, feature level, model parameter level, and model level, which correspond to the four ensemble forecasting methods that we have implemented based on the proposed SwinRNN and SwinVRNN models. 
The fixed distribution method can be applied during inference and is unnecessary for re-training the model, which is simple but gives unsatisfactory ensemble performance. 
In comparison, MC dropout is more effective and also simple to implement, and it can achieve better ensemble performance. 
MC dropout is just one of the possible model parameter perturbation methods, and other implementations are also worthy of further exploration.
SwinVRNN achieves the most desired ensemble performance among above perturbation methods with a single model. It improves the performance in almost all evaluated forecast times, even within the training period. Finally, multi-model SwinVRNN can further push upward the performance by simply assembling members from multiple SwinVRNN models.
However, there is also a limitation in SwinVRNN. When the perturbation module learns the parameters of the multivariate Gaussian distributions, the covariance between any two latent variables in different spatial locations need be calculated, which means a quadratic complexity of the image size.
The memory footprints would increase rapidly when dealing with high resolution inputs, such as ERA5 data with $0.25^\circ$ spatial resolution. There are two feasible methods to tackle the problem. One is to calculate the covariance matrix on features with lower spatial resolution, which is easy to implement but may lead to loss of some accuracy. The other is to calculate the covariance in a local region. Considering the sparsity of the learned covariance matrix (as shown in Figure \ref{fig-members} (b)) and the first law of geography, the correlation between variables in distant locations can be neglected.
Thus a more efficient and effective manner for modeling the multivariate normal distribution can be explored. 
One may also refer to the localization technique in Ensemble Kalman Filter (EnKF) for this problem. In addition, the combination of traditional NWP ensemble methods, such as singular vectors (SVs), and conditional nonlinear optimal perturbation (CNOP), with deep learning models is also worth being explored in future.



\section{Conclusions}
This paper has mainly focused on data-driven based ensemble forecasting. 

First, we proposed a novel stochastic weather prediction model --- SwinVRNN. It consists of a deterministic SwinRNN backbone and a perturbation module. We justified the effectiveness of the designs of SwinRNN, including the multi-scale decoder, residual prediction, and Swin attention based memory update mechanism. We then proposed the perturbation module to learn a stochastic latent variable following the Variational Auto-Encoder paradigm. The learned latent variable is leveraged to perturb the hidden features of the SwinRNN model. We verified the effectiveness of SwinVRNN on ensemble forecasting and the significance of considering the spatial covariance, via comparison experiments on WeatherBench dataset. 

In addition, we designed four categories of ensemble forecasting approaches based on our SwinRNN and SwinVRNN models, including fixed distribution perturbation, learned distribution perturbation, model parameters perturbation, and multi-model ensemble, and compared their ensemble performance on lead times up to 14 days. The learned distribution perturbation implemented by our SwinVRNN model outperforms fixed distribution and MC dropout methods significantly, and the multi-model ensemble of SwinVRNN further boosts the performance.

Finally, we compared the forecasting performance of our SwinRNN and SwinVRNN models with three categories of baselines, i.e. the weekly climatology, NWP baselines, and data-driven baselines. The proposed models set a new state-of-the-art among data-driven based weather forecasting models. More notably, our models outperform IFS on T2M and TP variables on all lead times up to 5 days.

\bibliographystyle{unsrt}  
\bibliography{references}

\end{document}